\pdfoutput=1

\documentclass[11pt]{article}

\newcommand\blfootnote[1]{%
  \begingroup
  \renewcommand\thefootnote{}\footnote{#1}%
  \addtocounter{footnote}{-1}%
  \endgroup
}

\usepackage[]{acl}

\usepackage{times}
\usepackage{latexsym}
\usepackage{comment}
\usepackage{color}
\usepackage{listings}

\usepackage{colortbl}
\usepackage[T1]{fontenc}

\usepackage[utf8]{inputenc}

\usepackage{microtype}
\usepackage[most]{tcolorbox}
\usepackage{inconsolata}

\usepackage{mathtools}
\usepackage{makecell}

\usepackage{array,multirow,graphicx}
\usepackage{amsmath}
\usepackage{amssymb}
\usepackage{booktabs}
\usepackage{algpseudocode}

\usepackage[export]{adjustbox}
\usepackage{float}

\usepackage[capitalize]{cleveref}
\crefname{section}{Sec.}{Secs.}
\Crefname{table}{Table}{Tables}

\usepackage{amsmath,amsfonts,bm}

\def\eqref#1{equation~\ref{#1}}

\def\1{\bm{1}}

\DeclareMathAlphabet{\mathsfit}{\encodingdefault}{\sfdefault}{m}{sl}
\SetMathAlphabet{\mathsfit}{bold}{\encodingdefault}{\sfdefault}{bx}{n}

\usepackage{geometry}
\usepackage{booktabs} 
\usepackage{bbm}
\usepackage{mathtools}
\usepackage{nccmath}
\usepackage{setspace}

\usepackage{caption}
\usepackage{subcaption}

\usepackage[linesnumbered,ruled,vlined]{algorithm2e}

\SetKwInput{KwInput}{Input}                
\SetKwInput{KwOutput}{Output}              

\SetCommentSty{mycommfont}

\SetAlCapSty{algcapsty}

\usepackage[T1]{fontenc}
\usepackage{wrapfig,lipsum,booktabs}

\usepackage{soul}
\usepackage{dsfont}
\usepackage{enumerate}
\usepackage{enumitem}

\usepackage{amsmath}
\usepackage{amsfonts}
\usepackage{bbm}
\usepackage{dsfont}
\usepackage[Symbol]{upgreek}
\usepackage{lscape}
\usepackage{caption}
\usepackage{balance}
\usepackage{xspace}
\usepackage{float}
\usepackage{kotex}

\usepackage{wasysym}
\usepackage{xcolor}
\usepackage{multirow}
\usepackage{array, boldline, rotating}

\usepackage{amssymb}
\usepackage{pifont}
\renewcommand*\eqref[1]{(\ref{#1})}

\newcommand{\yes}[1]{\textcolor{blue}{[YES]}}
\newcommand{\no}[1]{\textcolor{orange}{[NO]}}
\newcommand{\na}[1]{\textcolor{gray}{[N/A]}}

\newcommand{\ie}{\emph{i.e.,~}}

\definecolor{LightCyan}{rgb}{0.88,1,1}
\definecolor{Blue}{rgb}{0, 0.5, 1}
\definecolor{Green}{rgb}{0.0, 0.8, 0.0 }
\definecolor{Red}{rgb}{0.95, 0.55, 0.6}
\definecolor{Skyblue}{rgb}{0.6, 0.6, 0.95 }

\NewDocumentEnvironment{suptitle}{ +b }{
    \twocolumn[{#1}]%
}{}

\NewDocumentCommand{\supptitle}{s}{
\begin{suptitle}
        \centering
        \rule{\textwidth}{0.03cm}\\[0.1cm]
        - Appendix -\\[0.2cm]
        {\Large 
            \textbf{\mytitle }
        }\\
        \rule{\textwidth}{0.03cm}\\[0.2cm]
\end{suptitle}}

\newcommand{\mytitle}{Chain-of-Rank: Enhancing Large Language Models \\ for Domain-Specific RAG in Edge Device}
\title{\mytitle}

\author{Juntae Lee\hspace{1em}Jihwan Bang\hspace{1em}Seunghan Yang\hspace{1em}Kyuhong Shim\hspace{1em}Simyung Chang\\
{Qualcomm AI Research$^\dag$} \\ 
{\texttt {\small\{juntlee, jihwbang, seunghan, kshim, simychan\}@qti.qualcomm.com}}}

\begin{document}
\maketitle

\blfootnote{\hspace{-1.8em}$^\dag$Qualcomm AI Research is an initiative of Qualcomm Technologies, Inc. and/or its subsidiaries.}

\begin{abstract}
Retrieval-augmented generation (RAG) with large language models (LLMs) is especially valuable in specialized domains, where precision is critical. 
To more specialize the LLMs into a target domain, domain-specific RAG has recently been developed by allowing the LLM to access the target domain early via finetuning. 
The domain-specific RAG makes more sense in resource-constrained environments like edge devices, as they should perform a specific task (e.g. personalization) reliably using only small-scale LLMs.
While the domain-specific RAG is well-aligned with edge devices in this respect, it often relies on widely-used reasoning techniques like chain-of-thought (CoT). The reasoning step is useful to understand the given external knowledge, and yet it is computationally expensive and difficult for small-scale LLMs to learn it. 
Tackling this, we propose the Chain of Rank (CoR) which shifts the focus from intricate lengthy reasoning to simple ranking of the reliability of input external documents. Then, CoR reduces computational complexity while maintaining high accuracy, making it particularly suited for resource-constrained environments. We attain the state-of-the-art (SOTA) results in benchmarks, and analyze its efficacy.

\end{abstract}
\section{Introduction}

The integration of retrieval-augmented generation (RAG) with large language models (LLMs)~\cite{lewis2020retrieval} has emerged as a pivotal advancement in mitigating the issue of factual hallucination~\cite{ji2023survey}—an inherent limitation of LLMs when generating knowledge-intensive responses. By leveraging external knowledge sources, 
RAG enables LLMs to utilize relevant knowledge dynamically, enhancing both the accuracy and reliability of their outputs. 

RAG is especially crucial in the context of specialized domains, where precision is paramount and errors can be costly. Also, in RAG, LLMs must not only incorporate the relevant external information as the input, but also contextualize the information within the nuances of the target domain. To optimize the RAG-LLM for a specific domain, recently domain-specific RAG~\cite{RAFT} has been developed where LLMs can early access the target domain through finetuning. The practicality of the domain-specific RAG is more noteworthy when computational resources are limited such as edge devices since with only a small-scaled LLM some tasks should be performed reliably.
\begin{figure}[t]
    \centering
    \includegraphics[width=0.8\linewidth]{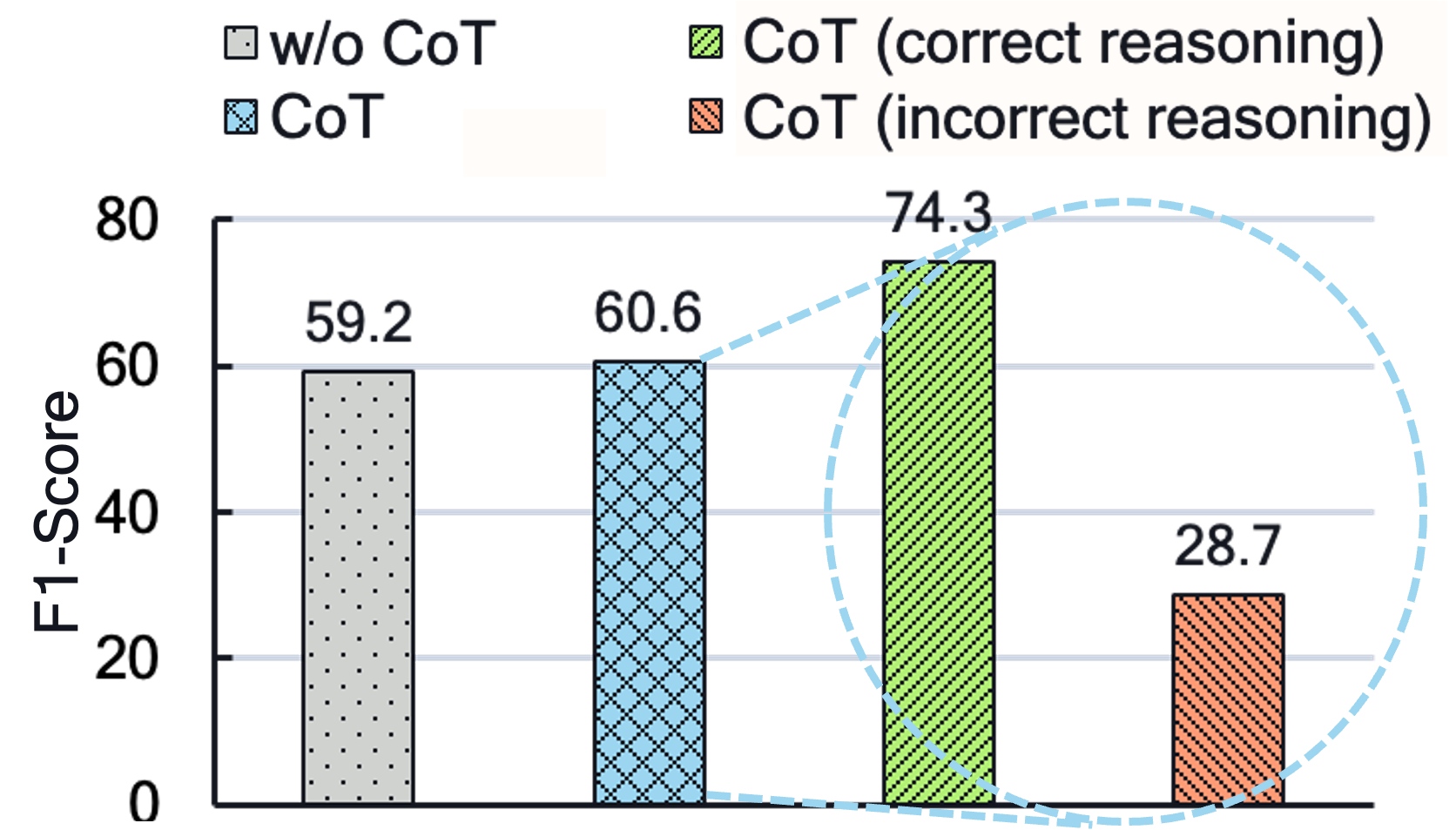}
    \caption{\textbf{Performance for domain-specific RAG on HotPotQA dataset on LLaMA3-8B with LoRA adapter.} The marginal effect of CoT (59.2\% $\rightarrow$ 60.6\%) is because of the generated incorrect reasoning which severely degrades the performance.}
    \label{fig:intro}
    \vspace{-0.2cm}
\end{figure}

Despite the promise of the domain-specific RAG, the input external knowledge (generally retrieved information dubbed contexts) may consist of both irrelevant and relevant contexts. 
Hence, reasoning process such as chain-of-thought (CoT)~\cite{wei2022chain} is useful for understanding and focusing on the relevant context. To this end, in RAFT~\cite{RAFT}, LLM learns the reasoning as well as answering in finetuning. Also, in~\cite{con}, the reasoning is prompted to generate a summary of all the contexts. Elaborated reasoning is beneficial, and yet obtaining this kind of reasoning dataset for domain-specific learning is time-consuming and costful, and also it incurs a large testing cost. 

\begin{figure*}[t]
    \centering
    \includegraphics[width=\linewidth]{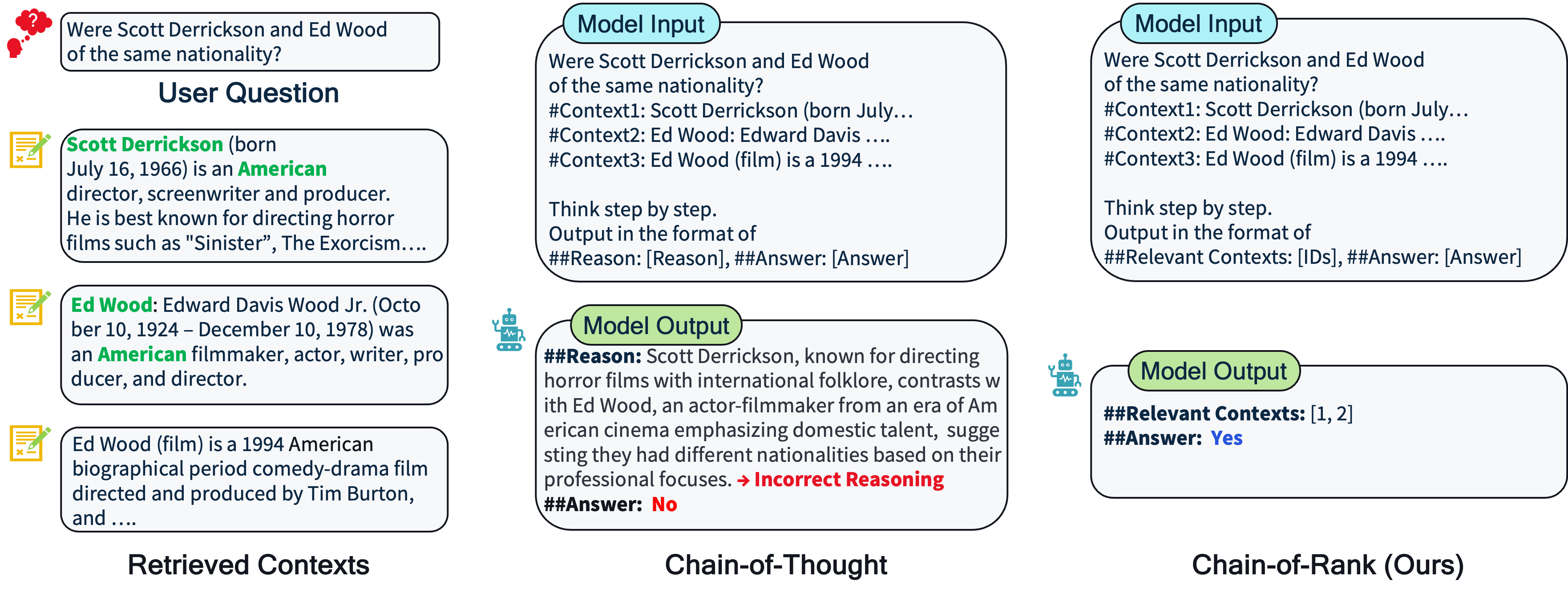}
    \vspace{-0.5cm}
    \caption{\textbf{Illustration of the proposed chain-of-rank for domain-specific RAG.} CoR streamlines the reasoning step, which is easier to be learned.}
    \label{fig:main}
    \vspace{-0.3cm}
\end{figure*}

Unreliability of reasoning is also a critical issue, especially when the parameter-efficient fine-tuning (PEFT) like LoRA adapters~\cite{hu2021lora,huang2023lorahub,bang2024crayon} are used to reduce the computational burden in training resource constraint environment. Namely, the PEFT adapters are efficient but lack enough learning capacities, and then struggle to learn the intricate reasoning process. As shown in Fig.~\ref{fig:intro}, LLaMA3-8B~\cite{dubey2024llama} with LoRA exhibits marginal gains when learned with CoT. It means that the intricate reasoning can become a hindrance~\cite{shi2023large} in resource-constrained domain-specific RAG.

Instead of focusing the intricate reasoning processes, we narrow the focus to the ranking of the contexts' relevance, then the LLM can streamline its reasoning and reduce the cognitive load required to generate accurate final answers. Building on this insight, we propose the Chain-of-Rank (CoR) approach. In this method, the model is learned to output just the ID of the contexts which are relevant to the query, and the answer. Then, CoR not only reduces computational complexity but also enables the LLM to concentrate more fully on the critical information, leading to more accurate and domain-specific outputs. This focus on relevance rather than elaborate reasoning aligns well with the resource limitations of small-scale LLMs and edge devices.

\section{Related Works}
\vspace{-0.2cm}
\noindent \textbf{Domain-specific RAG.}
In the existing training-based RAG~\cite{linra,wanginstructretro,asai2023self}, the LLM is learned for various domains, and then applied to unseen domains. However, for better contextualization or under constrained resource condition, it is beneficial for LLM to be early accessed to the target domain via training on the domain. To this end, RAFT~\cite{RAFT} pioneered domain-specific RAG. In RAFT, the LLM is learned by alternating two loss functions which are designed to simulate open-book and closed-book cases, respectively. The first loss addresses both distracting and golden contexts, while the second loss does only distracting ones. However, for decent performance evenly across various datasets, they trained the LLM to learn how to make the intricate reasoning as well as the answer.  

\noindent \textbf{Reasoning techniques in LLM.}
CoT~\cite{wei2022chain} reasoning has been shown to enhance performance in LLMs, sparking numerous studies aimed at improving its efficiency. To delve into CoT, more complex approaches like CoT decoding by sampling~\cite{wang2022self,wang2024chain} and analogous reasoning~\cite{yasunaga2023large} have emerged. Considering the lengthy inputs which contain the retrieved contexts as well as the query in RAG, sampling to find the optimal decoding path or generating reasoning examples by itself are too burden in terms of computational cost. Tailored for RAG, methods like RAFT~\cite{RAFT} and CoN~\cite{con} have demonstrated the effectiveness of reasoning in RAG. However, as highlighted in~\cite{shi2023large}, errors in reasoning can lead to incorrect answers. When using low-capability LLMs to learn both reasoning and answering, these errors become more pronounced. Since retrieved contexts contain factual knowledge, focusing on simpler and more efficient reasoning that just prioritizes relevant contexts can mitigate this issue, making the identification of relevant context alone sufficient.

\section{Method}

\subsection{RAG Problem Set-up}
In RAG, an LLM can be formalized as $p(y|x) = \sum_{D} p(y|x, D) p(D|x)$, where $x$ denotes the input query, $y$ represents the LLM generated answer, and $D=\{d^k\}_{i=1}^{K}$ contains the $K$ individual contexts. This formulation takes into account the joint probability of retrieving a set of contexts, rather than assuming the contexts are selected independently.

As this sum over all the context sets is impractical, generally an off-the-shelf retriever selects the top-$K$ most relevant contexts. This leads to the approximation: $p(y|x) \approx p(y| x, D)$. Furthermore, when a reasoning step is considered, it becomes as follows.
\begin{equation}
\label{eq1} 
    p(y|x) \approx \sum_{R} p(y|x, D, R) p(R|x,D)
\end{equation}
where $R$ represents the generated reasoning.

\subsection{Chain-of-Rank} 

\noindent\textbf{Framework.}
We streamline the reasoning process by shifting the focus from complex reasoning to identifying the IDs of the given contexts that correspond to the most relevant ones for $x$. With just this process, the model can reduce cognitive overhead on less relevant information, and more pay attention to the relevant information. As illustrated in Fig.~\ref{fig:main}, each context is identified by its unique ID. Then, CoR involves two main steps: (1) selecting the ID of relevant contexts (\ie $R$) and (2) generating the final answer $y$.
Consequently, the CoR framework significantly simplifies the reasoning step, making it a practical solution for scenarios with limited computational resources while enhancing performance in domain-specific applications.

\noindent\textbf{Model training.}
We concatenate the instruction, question, and retrieved documents into a single prompt, allowing the model to learn in a standard supervised manner. The model is trained to optimize both the selection of relevant document IDs and the accuracy of the generated answer by minimizing a joint loss function. 
\begin{equation}
\label{eq2} 
    \mathcal{L} = - \sum_{i=1} \log p(y_i | x_i, D_i, R_i) - \log p(R_i | x_i, D_i)
\end{equation}
We designed the top-$k$ documents in $D_i$ to include at least one positive document for a query $x_i$ during training. Also, we employed LoRA to efficiently fine-tune the model parameters assuming the low resource constraint (details are in Appendix).

\section{Experiments}
We provide more details and analysis in Appendix.

\noindent \textbf{Datasets.}
In our experiments, we use the following datasets to evaluate the proposed method. We selected these datasets to represent both popular and diverse domains including Wikipedia and Coding/API documents. 


\begin{table*}[t]
  \centering
  \begin{adjustbox}{width=0.8\linewidth}
  \begin{tabular}{@{}clccccc@{}}
    \toprule
    & \multirow{2}{*}{Method}      &      \multicolumn{2}{c}{HotPotQA}      &      \multirow{2}{*}{TensorFlow}    &      \multirow{2}{*}{HuggingFace}      &      \multirow{2}{*}{TorchHub} \\
    \cmidrule(ll){3-4}
    && EM& F1 score& & & \\ 
    \midrule
    & LLaMA3-8B & 40.84 & 52.47 & 32.11 & 10.14 & 22.13 \\
    \midrule
    \multirow{4}{*}{Domain-specific} & DSF & 44.98 & 59.15 & 83.91 & 87.42 & 70.08\\
    & RAFT (DSF-CoT) & 46.79 & 60.59 & 88.98 & 89.68 & 74.05\\
    & DSF-CoN & 48.60 & 62.04 & 84.52 & 79.05 & 76.21 \\
    \cmidrule(ll){2-7}
    & \textbf{DSF-CoR (Ours)} & \textbf{49.23} & \textbf{64.11} & \textbf{95.68} & \textbf{92.52} & \textbf{80.54} \\

    \bottomrule
    
  \end{tabular}
  \end{adjustbox}
  \caption{\textbf{Comparative results on domain-specific RAG.} EM and F1 score for the HotPotQA, and AST sub-tree matching accuracy scores (\%) for the Gorilla API (TensorFlow, HuggingFace, TorchHub) are reported.}
  \label{tab:main}
\end{table*}

\begin{table}[t]
  \centering
  \begin{adjustbox}{width=\linewidth}
  \begin{tabular}{@{}lccc@{}}
    \toprule
    Method & DSF-CoT & DSF-CoN & \textbf{DSF-CoR} \\
    \midrule
    Reasoning Accuracy ($\uparrow$) & 68.21 & 69.02 & 72.31 \\
    Reasoning Tokens ($\downarrow$) & 90.15 & 143.18 & 8.00\\
    \bottomrule
    
  \end{tabular}
  \end{adjustbox}
  \caption{\textbf{Analyses on reasoning.} Accuracy (\%) and cost (used tokens) for reasoning are on the HotPotQA.}
  \label{tab:analysis}
  \vspace{-0.5cm}
\end{table}

In specific, we select HotPotQA~\cite{yang2018Hotpotqa} and Gorilla API datasets~\cite{patil2023gorilla}. The HotPotQA is the open-domain question-answers based on Wikipedia, mainly focused on common knowledge. In testing, we used `HotpotQA distractor dev. set' which is designed to provide ten contexts including at least one relevant context for a query. TensorFlow, HuggingFace, and TorchHub of the Gorilla API are to measure how to generate the correct, functional, and executable API calls based on the documentation. For each of HuggingFace, TorchHub, and TensorFlow, train and test splits are provided, which share the API pool. Also, following the officially-released code\footnote{https://github.com/ShishirPatil/gorilla}, we utilized BM25 retriever.

\noindent \textbf{Evaluation.} We set $K$ as $10$ for the HotPotQA, then all the available irrelevant contexts distract the relevant ones for a query. For the Gorilla API, $K$ is set to $5$. To minimize the influence of the quality of the off-the-shelf retriever, we set up our experiments to include at least one relevant context for each input query. 
For the HotPotQA, we used two standard metrics: Exact Match (EM) and F1 score, following prior work~\cite{chen1704reading,karpukhin2020dense,zhu2021retrieving}. An answer is correct in the EM if its normalized form, based on~\citep{karpukhin2020dense}, matches exactly the ground-truth answer. F1 score calculates token overlap between the prediction and ground truth~\cite{zhu2021retrieving}. For the Gorilla API, following the official benchmark, we perform AST sub-tree matching to identify which API the LLM is calling by matching key arguments, and report AST accuracy.

\noindent \textbf{Baselines.} 
We consider the following baselines for our experiments based on LLaMA3-8B: Naive LLaMA3-8B, Domain-specific fine-tuning (DSF) without reasoning, RAFT (DSF with CoT)~\cite{RAFT}, DSF + CoN (chain-of-note)~\cite{con}. In the DSF baselines, we commonly used the LoRA adapter. And, all the baselines are in the zero-shot setting. In addition, RAFT suggested to alternating two losses of the irrelevant contexts-only and the mixing irrelevant and relevant contexts. For a fair comparison, we tried to find the optimal combination ratio of the two losses for this baseline.


\subsection{Comparative Results}
\vspace{-0.2cm}
We evaluate our CoR and demonstrate the efficacy in Table~\ref{tab:main}. The non-specified pre-trained LLM (LLaMA3-8B) shows severely degraded scores in the API datasets than in the natural questions of HotPotQA, which proves the requirements and importance of domain-specific RAG. The reasoning-based methods, RAFT and CoN, attain better results than DSF. However, in F1 score, the effect of CoT is marginal. Also, although the noting strategy of CoN is tailored for RAG, it sometimes shows lower performance than the straightforward DSF as well as CoT (see TensorFlow and Huggingface results). Whereas, we see that the proposed CoR consistently and significantly outperforms the baselines in all the datasets. It means that learning the complex reasoning can be a burden to the PEFT on the smaller-scale LLM, and thus simply identifying the IDs of the relevant contexts is more beneficial.
We also study the extension of the proposed CoR to domain-agnostic RAG in Appendix.

\subsection{Analysis}
\noindent \textbf{Reasoning quality.}
In RAG, the reasoning can be utilized to support the answer. Therefore, the quality of reasoning is also substantial, then we quantitatively compare the reasoning of CoT, CoN, and our CoR. We evaluate CoT and CoN using a pre-trained LLM (e.g. GPT) in a massive scale. Since CoT and CoR produce lengthy reasoning that incorporates details of the retrieved contexts which may lead to some errors in detail. Hence, to ensure a fair comparison, we prompt the LLM evaluator to assess whether the reasoning is related with the relevant golden context. Nevertheless, in the top row of Table~\ref{tab:analysis}, the proposed CoR attains clearly higher reasoning accuracy.

\noindent \textbf{Cost in reasoning.} We also evaluate the proposed method in terms of the cost for reasoning. To this end, we compare CoT, CoN, and our CoR according to the number of the tokens used for reasoning. As shown in the bottom row of Table~\ref{tab:analysis}, CoR uses significantly lower tokens for reasoning, which shows the efficiency of the proposed approach.

\noindent \textbf{Importance of correct ranking}: To see this, we obtain the answer giving incorrect ranking for DSF-CoR. Despite domain-specific learning, it yields severely degraded results, 24.20\% EM and 32.34\% F1 score.

\section{Conclusions}
\vspace{-0.25cm}
We proposed the Chain of Rank (CoR) to address the limitations of the existing intricate reasoning processes like chain-of-thought in training-based, domain-specific RAG. For domain-specific RAG training, annotation expense for the reasoning data is required. Also, especially in testing on smaller LLMs in resource-constrained environments, it poses challenges in terms of the accuracy as well as computational cost. We observed that the inaccurate reasoning adversely affect the quality of final answer. By shifting the focus from elaborate reasoning to a simplified ranking of the reliability of retrieved documents, CoR significantly reduced computational complexity while attaining higher accuracy. Our experimental results demonstrated that CoR achieves SOTA results on RAG benchmarks, confirming its effectiveness in improving the domain-specific RAG performance of small-scale LLMs. 

\section{Limitations} 
\label{sec:limitation}
This work acknowledges the significance of reasoning in domain-specific RAG models and presents an efficient approach that reduces the need for complex training data labeling and significantly lowers reasoning costs during testing. However, we did not thoroughly investigate whether the proposed method would be equally effective in more general RAG frameworks that do not rely on task-specific training. That said, preliminary results presented in the appendix indicate the potential for success in general RAG settings, suggesting that this area warrants deeper exploration in future work. Therefore, our findings provide a promising foundation for future research.

\section{Ethical Consideration}
In the field of domain-specific RAG, if the applications involve sensitive areas such as personal information, special caution must be taken during the model training process to ensure privacy and data protection. Beyond this consideration, methodologically, our research focuses on improving the accuracy and efficiency of RAG in LLMs, we do not foresee any direct negative ethical concerns stemming from our contributions. Nonetheless, it is important to recognize that generative AI technologies, including those using LLMs, come with potential risks. As such, careful consideration of their broader ethical and societal implications is necessary when these systems are applied in the real world.

\bibliography{bib}

\newpage
\appendix
\supptitle

\section{Prompt template for chain-of-rank}
\begin{tcolorbox}[fonttitle=\small\bfseries,
fontupper=\scriptsize\sffamily,
fontlower=\fon{put},
colback=gray!5!white, colframe=gray!75!black,
enhanced,
left=2pt, right=2pt, top=2pt, bottom=2pt,
title=Prompt template for domain-specific RAG with CoR]
\begin{lstlisting}[]
Contexts and Question are given. 

Let's think step by step to make 
correct output. 

First, reranking goal: select the 
relevant contexts, important to answer 
the question correctly. 

Then, answering goal: Focusing on the 
selected context, answer the question. 


Question: {question}
Context1: {context_1}
Context2: {context_2}
...
ContextN: {context_N}

Output:
## Relevant Context ID: {IDs}
## Answer: {answer}
\end{lstlisting}
\end{tcolorbox}

\section{Feasibility in task-agnostic RAG-LLM.} 
To identify the potential of the proposed chain-of-rank as the general reasoning technique in RAG beyond domain-specific RAG, we applied the proposed method on the pre-trained model (LLaMA3-8B). As shown in the Table~\ref{tab:agnostic}, the CoR is comparable to CoT.
Further, we combined the reasoning of the CoT and CoR. On top of the CoT-style reasoning, the reasoning of CoR makes meaningful synergistic results using a small cost.

\section{Datasets}
\noindent \textbf{HotPotQA.} HotPotQA is a large-scale question-answering dataset designed to evaluate both factual reasoning and multi-hop question answering. The training set with approximately 90,000 examples and development (dev.) set containing around 7,400 examples. For each question, ten contexts are provided where a context consists of several sentences and the key sentences (supporting facts to the query) are annotated. We experimented with the whole sentences in every context for a challenging set-up.

\begin{table}[t]
  \centering
  \begin{adjustbox}{width=0.8\linewidth}
  \begin{tabular}{@{}lcc@{}}
    \toprule
    Method & EM & F1 Score \\
    \midrule
    LLaMA3-8B pre-trained &  40.84 & 52.47 \\
    \; + CoT & 42.41 & 53.96\\
    \; + CoR & 41.20 & 55.92 \\
    \; + CoR\&CoT & 44.15& 58.09 \\
    \bottomrule
    
  \end{tabular}
  \end{adjustbox}
  \caption{\textbf{Results on a pre-trained LLM by applying the CoT, the proposed CoR, the mixture of CoT and CoR.} EM and F1 score are reported on the HotPotQA dataset.}
  \label{tab:agnostic}
\end{table}

\noindent \textbf{Gorilla API.} Gorilla API is multi-faceted, comprising three domains: TensorFlow, HuggingFace, TorchHub where training data includes 6190, 8191, 337 entries and testing data does 688, 911, 186 entries, respectively. Each entry of a domain conveys a detailed description for an API call. In specific, it consists of the following fields: \{domain, framework, functionality, api\_name, api\_call, api\_arguments, environment\_requirements, example\_code, performance, and description\}.

\noindent \textbf{Reasoning dataset for baseline training.}
Gorilla API dataset provides the explanation for every API document, and hence we use that as the reasoning following \cite{RAFT} for domain-specific training. In HotPotQA dataset which does not include reasoning, we utilized a significant-scale LLM to generate the intricate reasoning dataset for domain-specific training. We used the prompt of~\cite{RAFT} to generate the reasoning.


\section{Evaluation Metric}
\noindent \textbf{Exact Match.} Exact Match (EM) evaluates whether the model’s generated response is identical to the ground truth answer. It is computed as the percentage of predictions where the generated output exactly matches the reference answer, including the order and wording. EM is strict, meaning any deviation results in a score of 0 for that prediction, and only exact matches count as correct.

\noindent \textbf{F1 score.} F1 score is a measure that balances precision and recall. It is computed by comparing the overlap of tokens between the generated response and the ground truth. Precision is the ratio of correct tokens in the generated response to the total number of tokens in the response, while recall is the ratio of correct tokens to the total number of tokens in the ground truth. The F1 score is the harmonic mean of precision and recall, allowing for partial credit when the generated answer partially overlaps with the correct answer.

\noindent \textbf{AST accuracy}. AST (Abstract Syntax Tree) accuracy is a metric used to evaluate the correctness of generated API calls by comparing their structural representation to reference APIs. The generated API call is parsed into an AST, and its structure is matched against the corresponding reference API from the dataset. The accuracy is determined by how well the generated API’s function names and key arguments align with the reference. If the AST of the generated call matches a subtree of the reference API, it is considered correct.

\section{Prompt template to evaluate the reasoning}

\begin{tcolorbox}[fonttitle=\small\bfseries,
fontupper=\scriptsize\sffamily,
fontlower=\fon{put},
colback=gray!5!white, colframe=gray!75!black,
enhanced,
left=2pt, right=2pt, top=2pt, bottom=2pt,
title=Prompt template to evaluate reasoning]
\begin{lstlisting}[]
You are an expert at evaluating 
reasoning based on provided 
information. Given a question, five 
retrieved contexts, and reasoning, 
your task is to determine whether 
the reasoning is based on the correct 
context. The correct context is the 
one that contains the most relevant 
and accurate information to answer 
the question.

Follow these steps:
1. Identify which retrieved context 
contains the most accurate information 
to answer the question (the "golden 
context").
2. Evaluate if the reasoning is based 
primarily on this golden context.
3. Provide a clear answer (Yes or No).

### Question:
{question}

### Retrieved Contexts:
1. {context_1}
2. {context_2}
3. {context_3}
4. {context_4}
5. {context_5}

### Reasoning:
{reasoning}

### Evaluation:
Is the reasoning based on the correct 
context? Answer with "Yes" or "No".
\end{lstlisting}
\end{tcolorbox}

\section{LoRA-based training details}
We implemented the proposed and baseline approaches based on the Huggingface PEFT library~\cite{peft}. We set the rank $r$ and scaling factor of a LoRA as 128 and 16, respectively. In training, we use the AdamW optimizer with a learning rate 0.0003 which is cosine annealed. We also set the batch size as 128 and the maximum epoch as 1. All the proposed and baseline methods are implemented with PyTorch 2.0.1 and executed on two NVIDIA A5000 GPUs.

\end{document}